\documentclass{article}
\usepackage{ijcai24}

\usepackage[a-2b]{pdfx}
\usepackage{times}
\usepackage{soul}
\usepackage{url}
\usepackage[utf8]{inputenc}
\usepackage[small]{caption}
\usepackage{graphicx}
\usepackage{amsmath}
\usepackage{amsthm}
\usepackage{booktabs}
\usepackage{algorithm}
\usepackage{algorithmic}
\usepackage[switch]{lineno}

\usepackage{amssymb}
\usepackage{multirow}
\usepackage{subcaption}
\usepackage{makecell}
\usepackage{float}

\hypersetup{colorlinks=true, linkcolor=black, urlcolor=magenta, citecolor=black, filecolor=black}

\usepackage{xcolor}
\definecolor{lightgray}{RGB}{217, 217, 217}
\definecolor{forestgreen}{RGB}{47, 159, 87}%
\definecolor{forestred}{RGB}{202,12,22}%

\title{Revisiting Neural Networks for Continual Learning: An Architectural Perspective}

\author{
Aojun Lu$^1$
\and
Tao Feng$^2$\and
Hangjie Yuan$^3$\and
Xiaotian Song$^1$\And
Yanan Sun$^1$\footnote{Corresponding author.}
\affiliations
$^1$Sichuan University \\
$^2$Tsinghua University \\
$^3$Zhejiang University
\emails
aojunlu@stu.scu.edu.cn, 
fengtao.hi@gmail.com, 
hj.yuan@zju.edu.cn \\
songxt@stu.scu.edu.cn, 
ysun@scu.edu.cn
}

\begin{document}

\maketitle

\begin{abstract}

Efforts to overcome catastrophic forgetting have primarily centered around developing more effective Continual Learning (CL) methods. 
In contrast, less attention was devoted to analyzing the role of network architecture design (e.g., network depth, width, and components) in contributing to CL. 
This paper seeks to bridge this gap between network architecture design and CL, and to present a holistic study on the impact of network architectures on CL. 
This work considers architecture design at the network scaling level, i.e., width and depth, and also at the network components, i.e., skip connections, global pooling layers, and down-sampling. 
In both cases, we first derive insights through systematically exploring how architectural designs affect CL. 
Then, grounded in these insights, we craft a specialized search space for CL and further propose a simple yet effective \textbf{ArchCraft} method to steer a CL-friendly architecture, namely, this method recrafts \textit{AlexNet/ResNet} into \textit{AlexAC/ResAC}. 
Experimental validation across various CL settings and scenarios demonstrates that improved architectures are parameter-efficient, achieving state-of-the-art performance of CL while being \textbf{86\%}, \textbf{61\%}, and \textbf{97\%} more compact in terms of parameters than the naive CL architecture in \textit{Task IL} and \textit{Class IL}. Code is available at \url{https://github.com/byyx666/ArchCraft}.

\end{abstract}

\section{Introduction}
\label{sec:intro}

\begin{figure}[t]
\centering
\centerline{\includegraphics[width=0.98\columnwidth]{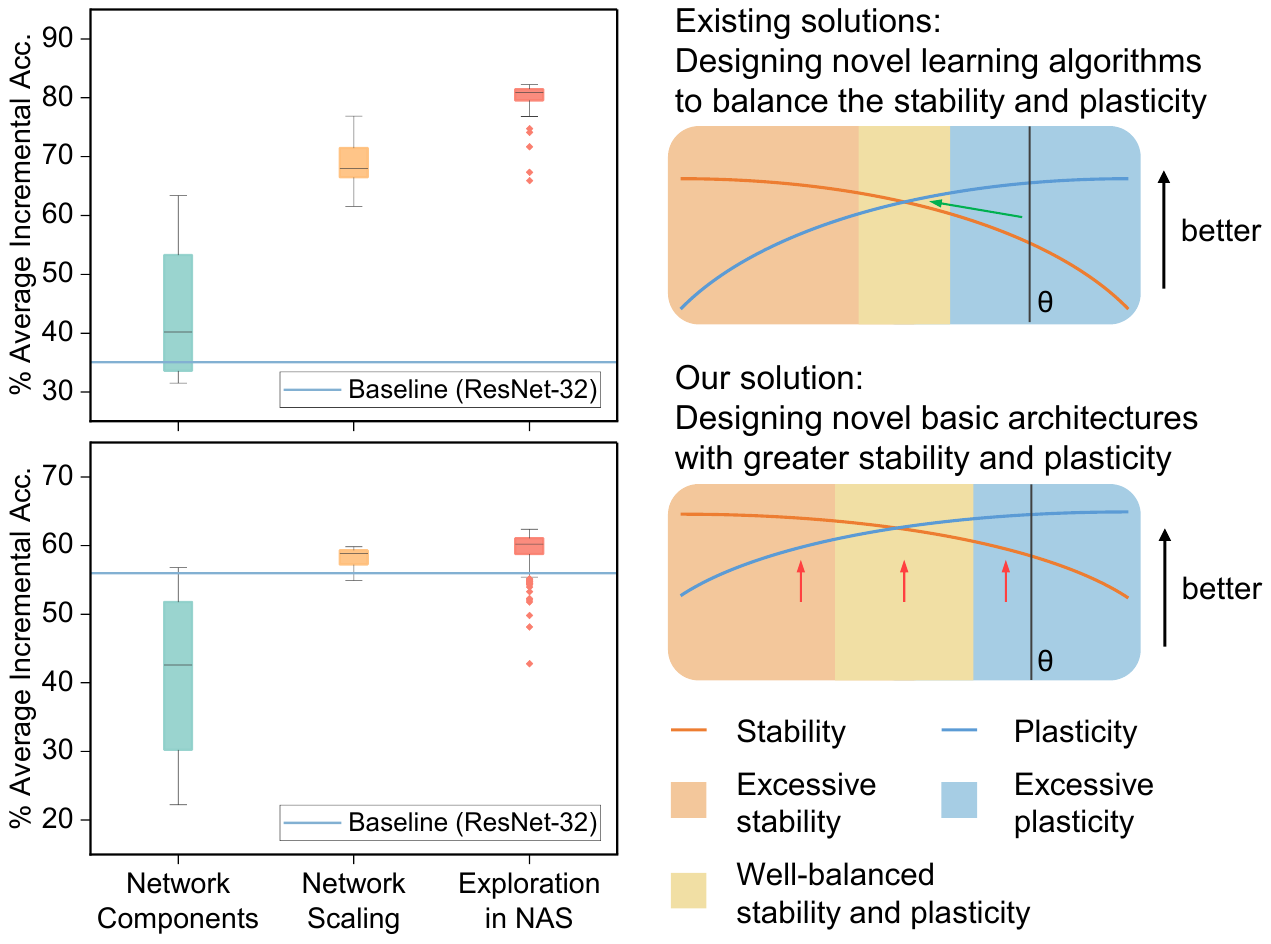}}
\vspace{-2pt}
\caption{\textbf{(\textit{Left}) Impact of architectural designs on CL} in \textit{Task IL} (top) and \textit{Class IL} (bottom). This displays the distributions of the CL performance with variations in network components and scaling. 
\textbf{
(\textit{Right}) Illustration of \textbf{ArchCraft}} for improving CL performance. 
Existing methods aim to seek the optimal balance of stability and plasticity (\textcolor{forestgreen}{green} arrow), ArchCraft focuses on enhancing stability and plasticity by recrafting the basic architecture (\textcolor{red}{red} arrows).
}
\label{fig:fig1}
\vspace{-6pt}
\end{figure}

Artificial Intelligence (AI) is expected to feature a robust capability to continuously acquire and update knowledge like creatures. To achieve this, Continual Learning (CL), also known as Incremental Learning (IL), has garnered much attention in AI~\cite{wang2023comprehensive}. Early, some efforts~\cite{li2017learning,serra2018overcoming} centered on Task Incremental Learning (\textit{Task IL}), where the task identity is given during both training and inference phases to assign a classifier. Recently, more works~\cite{feng2022overcoming,chen2023dynamic,bian2024make} focused on the more intricate Class Incremental Learning (\textit{Class IL}), in which the task identity is accessible only during training.

In both \textit{Task IL} and \textit{Class IL}, it is essential for the model to possess the plasticity required to acquire new knowledge and the stability necessary to retain previously learned knowledge.
However, the stability-plasticity dilemma remains a persistent challenge in CL~\cite{grossberg2013adaptive}. 
To address this, existing works~\cite{kirkpatrick2017overcoming,rebuffi2017icarl,shibata2021learning,wang2022foster} focus on developing novel algorithms to trade-off stability and plasticity. Among them, expansion-based methods dedicate different incremental network architectures to minimize conflicts between stability and plasticity. For instance, DER~\cite{yan2021dynamically} centers on feature extractor scaling, by contrast, MEMO~\cite{zhou2022model} puts effort into specific layers. In essence, these efforts resort to network width to build a good CL model. Moreover, some works can be formulated as utilizing network components to resist forgetting, e.g., CN~\cite{pham2022continual}, FPF~\cite{zhao2023does}, and LoRA~\cite{2023arXiv230906256L}.
However, these methods are built upon existing network architectures, which limits the scope of architecture exploration for better stability and plasticity.

A clear trend is that expanded architectures outperform the original ones. This raises a concern about whether the existing basic architectures of the network are ideal for CL. 
As shown in Figure~\ref{fig:fig1}, we observe that (i) various architectural designs significantly shape the impact towards CL, and (ii) directly using the native ResNet~\cite{he2016deep} leads to performance far from satisfactory. 
This motivates us to design architectures tailored for CL.

\textbf{How to seek an efficient and CL-friendly architecture?} Naturally, Neural Architecture Search (NAS)~\cite{elsken2019neural} is a straightforward candidate for exploring architectures for CL due to the automated search that can be performed on architectures. In this paper, we employ NAS as a promising solution for CL and probe catastrophic forgetting from an architectural perspective. However, the performance of NAS hinges on the quality of the search space design~\cite{radosavovic2020designing}. Given the limited prior knowledge regarding the influence of architectural designs on the performance of CL, constructing a suitable search space remains challenging. Thus, it becomes imperative to empirically glean this prior knowledge before employing NAS.

In short, this paper aims to bridge the gap between network architectures and CL. To this end, we (i) systematically scrutinize the influence of architectural design on CL, (ii) identify critical design candidates that can boost CL, and (iii) propose a NAS method with a novel search space tailored for CL.

Given that ResNet stands as a cornerstone in the AI era, we initiate this work from it to revisit neural networks and assess the impact of architectural designs on CL. Consequently, we formulated this study around network scaling and network components. Concerning the former, we consider the network width and depth. Regarding the latter, we delve into the investigation of skip connections, global pooling layers, and down-sampling approaches. Subsequently, grounded in empirical observations, we craft a specialized search space for CL, each candidate architecture within this space is succinctly represented, facilitating an efficient search process. In conclusion, we propose a simple yet effective method to \textbf{Craft} CL-friendly \textbf{Arch}itectures, dubbed \textbf{ArchCraft} (\textit{AC}). ArchCraft recrafts \textit{AlexNet/ResNet} into \textit{AlexAC/ResAC} to guide a well-designed network architecture for CL.
As a result, ArchCraft achieves superior CL performance while employing significantly fewer parameters than the original architecture.

To sum up, the contributions of this work are as follows:

\begin{itemize}
\setlength\itemsep{-0.1em}

\item We thoroughly scrutinize the mechanisms through which neural architectural design affects CL, and demonstrate that wider and shallower architectures better align with an effective CL model.

\item To the best of our knowledge, this paper is the pioneering effort to employ neural architecture design to shape a CL-friendly architecture. To achieve this, we propose a novel ArchCraft method to steer an architecture with greater stability and plasticity.

\item Extensive experiments show that the proposed method is parameter-efficient across various CL scenarios, i.e., achieving state-of-the-art performance with much fewer parameters than the baseline. Furthermore, our method features a controllable parameter.

\end{itemize}


\section{Related work}
\label{sec:rw}

\subsection{Continual Learning and Neural Architectures}

\paragraph{Continual Learning.} Neural networks have achieved remarkable success in various fields~\cite{yuan2023rlipv2,WinGNN}, but are ill-equipped for CL due to catastrophic forgetting. A series of works, such as encompassing regularization, dedicated memory systems, and modular architectures, are proposed to overcome forgetting. In detail, Regularization-based methods regularize the variation of network parameters or logit outputs to maintain stability, e.g., EWC~\cite{kirkpatrick2017overcoming} and XK-FAC~\cite{lee2020continual}. Replay-based methods preserve a few exemplars from previous tasks and replay them in learning a new task to prevent forgetting, e.g., iCaRL~\cite{rebuffi2017icarl} and Gcr~\cite{tiwari2022gcr}. This category of work designs different sampling strategies to establish the limited memory buffer for replay. Expansion-based methods continually expand the network methods, in which different incremental parts of the networks are dedicated to specific tasks, e.g., BNS~\cite{qin2021bns} and MEMO~\cite{zhou2022model}. This category of work focuses on the efficient utilization of network architectures. 

\paragraph{Neural Architectures for CL.} In principle, the performance of neural networks is mainly decided by their weights and architectures. The CL methods mentioned above are mainly for the weights, while the research on neural architectures for enhancing CL is still in the infant. In the literature, the work~\cite{mirzadeh2022wide} concluded that wider network architectures suffer less catastrophic forgetting in \textit{Task IL}. Upon this, another work~\cite{mirzadeh2022architecture} further investigated the impact of certain network components on \textit{Task IL}. To the best of our knowledge, there is no systematic work dedicated to designing the CL-friendly basic architectures (i.e., the backbones).

To sum up, unlike these methods that focus on expanding or modular the specific architecture to improve CL, this work centers on the architecture itself. Notably, we extend the scope of architectural design to encompass a broader scenario, a.k.a more intricate \textit{Class IL}. And we further propose an effective method for designing enhanced network architectures tailored for CL.

\begin{table*}[t]
\centering
\resizebox{.94\textwidth}{!}{
\begin{tabular}{@{\hspace{2mm}}ccc|cccc|cccc@{\hspace{2mm}}}
    \toprule
    \multicolumn{3}{c|}{Network Components} &\multicolumn{4}{c|}{Performance in Task IL} &\multicolumn{4}{c}{Performance in Class IL}\\
    \cmidrule(lr){1-3} \cmidrule(lr){4-7} \cmidrule(lr){8-11}
    Down. &Skip &GAP &R32-LA &R32-AIA &R18-LA &R18-AIA &R32-LA &R32-AIA &R18-LA &R18-AIA\\
    
    \midrule
    \multirow{4}*{\makecell[c]{Strided\\Conv.}} &\checkmark &\checkmark 
    &25.80{\scriptsize±1.80} &35.06{\scriptsize±0.88} &38.12{\scriptsize±2.85} &49.68{\scriptsize±0.91} 
    &37.92{\scriptsize±0.64} &55.97{\scriptsize±0.85} &40.34{\scriptsize±0.95} &57.79{\scriptsize±1.09}\\
    &\checkmark	&×	
    &55.52{\scriptsize±1.41} &60.89{\scriptsize±1.83} &62.95{\scriptsize±3.45} &65.41{\scriptsize±1.92} 
    &22.63{\scriptsize±3.36} &27.37{\scriptsize±11.18} &38.85{\scriptsize±1.60} &50.29{\scriptsize±3.01}\\
    &× &\checkmark 
    &25.29{\scriptsize±1.48} &31.74{\scriptsize±0.90} &30.33{\scriptsize±2.28}	&41.75{\scriptsize±0.89} 
    &30.74{\scriptsize±2.11} &46.99{\scriptsize±1.48} &38.45{\scriptsize±0.66} &56.86{\scriptsize±0.82}\\
    &× &× 
    &38.30{\scriptsize±1.17} &45.58{\scriptsize±1.76} &57.66{\scriptsize±1.72} &62.13{\scriptsize±1.02} 
    &27.63{\scriptsize±2.79} &38.66{\scriptsize±6.24} &33.99{\scriptsize±3.77} &45.98{\scriptsize±4.53}\\
    
    \midrule
    \multirow{4}*{\makecell[c]{Max\\Pooling}} &\checkmark &\checkmark 
    &25.91{\scriptsize±2.35}	&35.58{\scriptsize±0.78} &39.89{\scriptsize±1.80}	&53.15{\scriptsize±0.69} 
    &\textbf{38.27}{\scriptsize±0.88} &\textbf{56.79}{\scriptsize±0.88} &\textbf{40.50}{\scriptsize±0.34} &\textbf{59.53}{\scriptsize±1.26}\\
    &\checkmark	&×	
    &\textbf{57.31}{\scriptsize±1.56} &\textbf{63.35}{\scriptsize±1.65} &\textbf{63.94}{\scriptsize±3.13} &\textbf{68.74}{\scriptsize±3.27} 
    &24.91{\scriptsize±3.58} &22.23{\scriptsize±13.19} &40.00{\scriptsize±0.67} &54.11{\scriptsize±1.79}\\
    &× &\checkmark 
    &24.54{\scriptsize±0.53}	&32.16{\scriptsize±0.87} &30.46{\scriptsize±2.52}	&42.24{\scriptsize±1.08}
    &30.69{\scriptsize±1.27} &47.50{\scriptsize±1.42} &37.34{\scriptsize±0.72} &56.53{\scriptsize±0.83}\\
    &× &× 
    &36.53{\scriptsize±0.85} &44.72{\scriptsize±1.31} &60.16{\scriptsize±0.88} &65.41{\scriptsize±0.69} 
    &16.67{\scriptsize±13.00} &25.64{\scriptsize±18.06} &33.92{\scriptsize±3.98} &47.83{\scriptsize±4.96}\\
    
    \midrule
    \multirow{4}*{\makecell[c]{Avg\\Pooling}} &\checkmark &\checkmark 
    &24.37{\scriptsize±1.72} &35.00{\scriptsize±0.92} &38.47{\scriptsize±2.01} &53.52{\scriptsize±0.67} 
    &38.25{\scriptsize±0.52} &56.67{\scriptsize±0.61} &39.85{\scriptsize±1.52} &57.53{\scriptsize±1.14}\\
    &\checkmark	&×	
    &56.16{\scriptsize±2.93} &62.60{\scriptsize±1.74} &63.92{\scriptsize±3.22} &68.56{\scriptsize±2.78} 
    &27.62{\scriptsize±5.27} &34.69{\scriptsize±11.64} &37.35{\scriptsize±3.96} &49.74{\scriptsize±4.24}\\
    &× &\checkmark 
    &23.70{\scriptsize±1.92} &31.50{\scriptsize±0.88} &30.21{\scriptsize±1.27} &42.82{\scriptsize±0.80} 
    &30.33{\scriptsize±2.09} &46.55{\scriptsize±1.24} &37.02{\scriptsize±0.92} &55.99{\scriptsize±0.48}\\
    &× &× 
    &35.86{\scriptsize±2.09} &44.93{\scriptsize±2.42} &60.68{\scriptsize±0.86} &65.78{\scriptsize±1.29} 
    &23.04{\scriptsize±11.10} &33.05{\scriptsize±14.94} &34.84{\scriptsize±1.67} &47.63{\scriptsize±1.41}\\
    
    \bottomrule
\end{tabular}
}
\vspace{-2pt}
\caption{The CL performance of the networks with different configurations of down-sampling approaches (denoted as `Down.') and whether to use skip connections (denoted as `Skip') and GAP or not. `R18' and `R32' represent networks based on ResNet-18 and ResNet-32.}
\label{tab:network_components} 
\vspace{-6pt}
\end{table*}

\subsection{Neural Architecture Search (NAS)}

NAS automates the design of high-performance neural architectures by formulating the design process as an optimization problem~\cite{elsken2019neural}. In this process, NAS employs an optimization method (i.e., search strategy), to traverse a predefined search space comprising candidate architectures. After searching, the architecture demonstrating the best performance is chosen as the final design. Some works have already empirically shown that NAS can craft architectures that surpass those manually designed~\cite{zoph2018learning}.

\textbf{Discussion.} NAS can be employed to design architectures for CL.    However, the crux of NAS lies in defining a suitable search space~\cite{radosavovic2020designing,wan2022redundancy}, which delineates a searchable subset of potential architectures from the vast architecture space. While numerous architectural design practices offer substantial prior knowledge about architectures with remarkable plasticity for standard learning paradigms~\cite{howard2019searching,tan2021efficientnetv2}, designing a search space specifically tailored for CL requires revisiting existing architectural design experiences. This is essential as architectures for CL need to possess not only plasticity but also stability~\cite{grossberg2013adaptive}, necessitating the construction of a CL-friendly search space.


\section{Preliminaries}	
\label{sec:setup}

In this section, we describe the experimental setup for training and evaluating the networks toward CL.

\textbf{Benchmark.} For the CL scenarios mentioned above, i.e., \textit{Task IL} and \textit{Class IL}, we assess network performance on CIFAR-100. The training process involves gradually introducing all 100 classes with 5 classes per incremental step, totaling 20 steps or tasks. In \textit{Task IL}, the network is trained using a vanilla SGD optimizer, while in \textit{Class IL}, a replay buffer containing 2,000 examples is employed.

\textbf{Evaluation Metrics.} Let $K$ be the number of tasks, the average classification accuracy after learning the $b$-th task, say $A_{b}$, is defined as:
\begin{equation}
    A_{b}=\frac{1}{b} \sum_{i=1}^{b} a_{i,b}
\end{equation}
where $a_{i,b}$ is the classification accuracy evaluated on the test set of the $i$-th task after learning the $b$-th task ($i \le b$). The performance of CL is measured by two metrics: the {\em Last Accuracy} (LA) and the {\em Average Incremental Accuracy} (AIA). The LA is the classification accuracy after the last task, i.e., $LA=A_{K}$, which reflects the overall accuracy among all classes. Further, the AIA denotes the average of $A_{b}$ over all tasks, i.e., $AIA=\frac{1}{K} \sum_{b=1}^{K} A_{b}$, which reflects the performance of all incremental stages. The higher LA and AIA, the better CL performance.

\textbf{Implementation Details.} In all experiments, we report the mean and standard deviation calculated across 5 runs with different random seeds. For \textit{Task IL}, we train the model by 60 epochs in the first task and 20 epochs in the subsequent tasks. For \textit{Class IL}, we follow PyCIL~\cite{zhou2023pycil} to train the model by 200 epochs in the first task and 70 epochs in the subsequent tasks.


\section{Design of CL-Friendly Architectures}
\label{sec:im_de}

In this section, we first decompose and study the architectural design on the performance of \textit{Task IL} and \textit{Class IL} at the network components and scaling. Upon these insights, we propose an ArchCraft method to craft CL-friendly architectures.

\subsection{Impact of Network Components}
\label{subsec:network_components} 

Strided convolution, skip connections, and Global Average Pooling (GAP) have all proven effective in neural networks in standard learning paradigms~\cite{he2016deep}. However, the impact of these components on CL remains elusive. To address this gap, we perform experiments to determine the optimal configurations of these components for both \textit{Task IL} and \textit{Class IL}. To be specific, the candidate configurations can be obtained by replacing strided convolution with average/max pooling, removing skip connections, or removing GAP. In particular, we utilize the ResNet series as the architectural framework and examine the impact of the aforementioned components on CL performance.

As shown in Table~\ref{tab:network_components}, different configurations of network components influence the performance of both \textit{Task IL} and \textit{Class IL}. In brief, we observe that the optimal configuration involves employing max pooling and skip connections while removing GAP for \textit{Task IL}. Moreover, the optimal configuration for \textit{Class IL} is similar to \textit{Task IL} but retains GAP. In particular, skip connections significantly improve the performance in both \textit{Task IL} and \textit{Class IL}, mirroring their role in standard learning paradigms. In contrast, the function of GAP diverges significantly between \textit{Task IL} and \textit{Class IL}. In \textit{Task IL}, eliminating the GAP in networks considerably enhances CL performance, while in \textit{Class IL}, it causes severe performance degradation.

\subsection{Impact of Network Scaling}
\label{subsec:network_scaling}

Width and depth are two crucial factors that affect the capacity and complexity of neural networks. We conduct a series of experiments on networks with different widths and depths to investigate the impact on CL performance. Specifically, we adopt ResNet-32 as the skeleton of the network~\cite{he2016deep}, which consists of a single convolution layer and three subsequent stages, each of them containing the same number of blocks. In that case, the width is represented by the initial number of channels, denoted by \textit{W}, and the depth is represented by the number of blocks in each stage, denoted by \textit{D}. Moreover, we note that all of the network architectures are modified by applying the optimal configurations of network components summarized in Section~\ref{subsec:network_components}.

\begin{figure}[ht]
    \centering
    \vspace{-2pt}
    \subcaptionbox{LA in Task IL\label{subfig:la_til}}{\includegraphics[width = 0.23\textwidth]{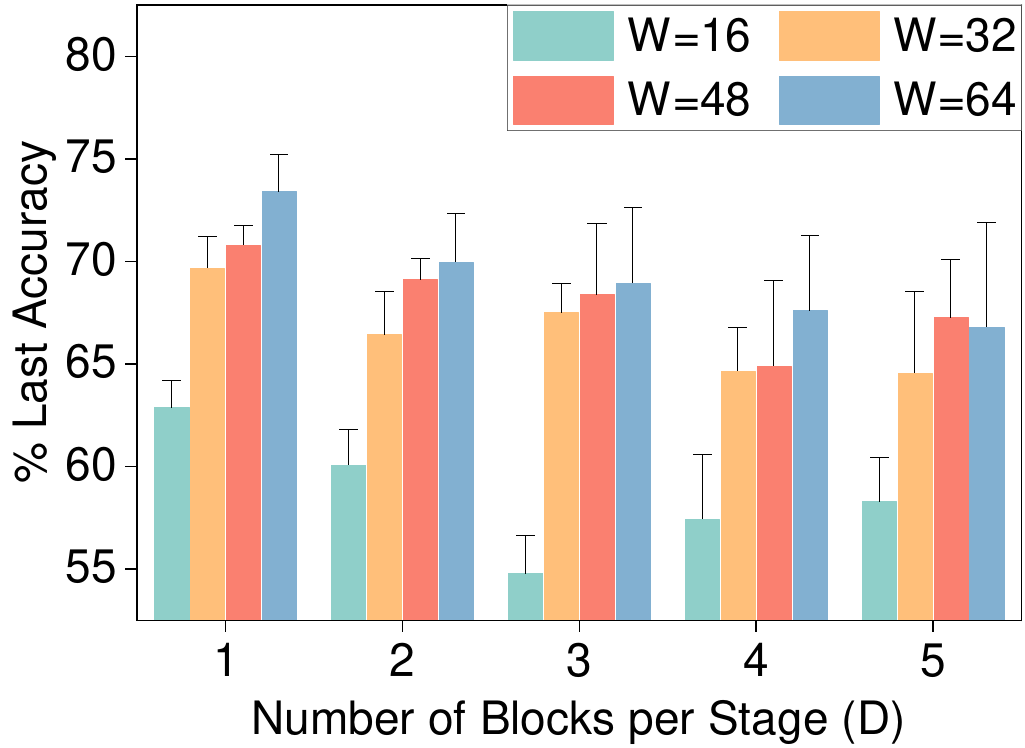}}
    \hfill
    \subcaptionbox{AIA in Task IL\label{subfig:aia_til}}{\includegraphics[width = 0.23\textwidth]{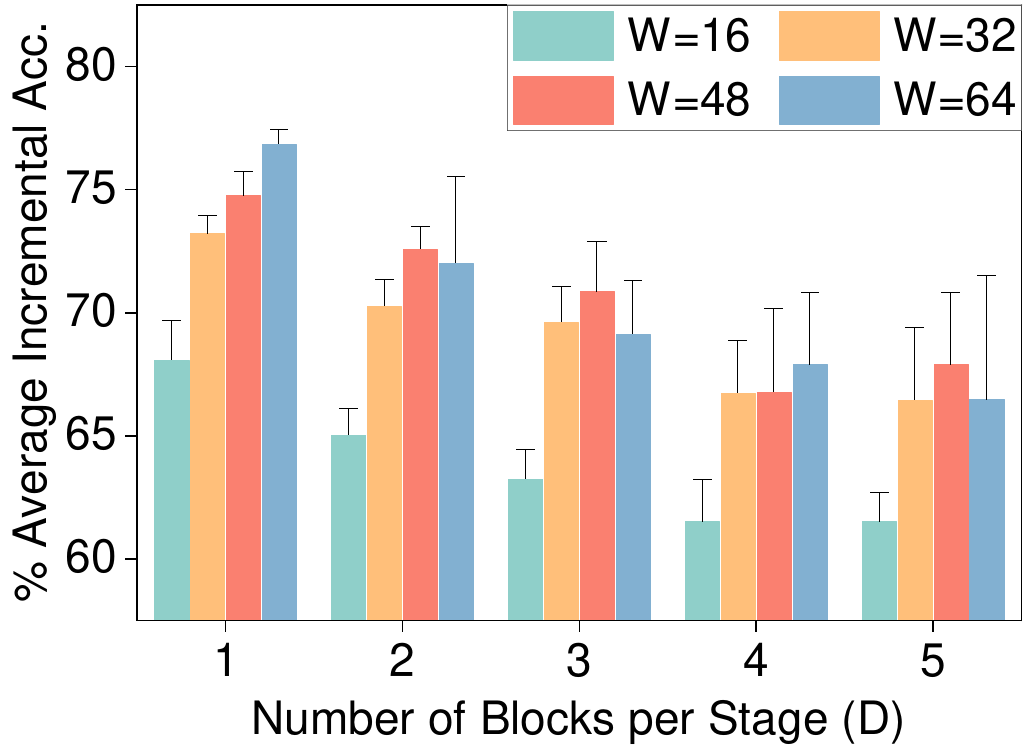}}
    \\
    \subcaptionbox{LA in Class IL\label{subfig:la_cil}}{\includegraphics[width = 0.23\textwidth]{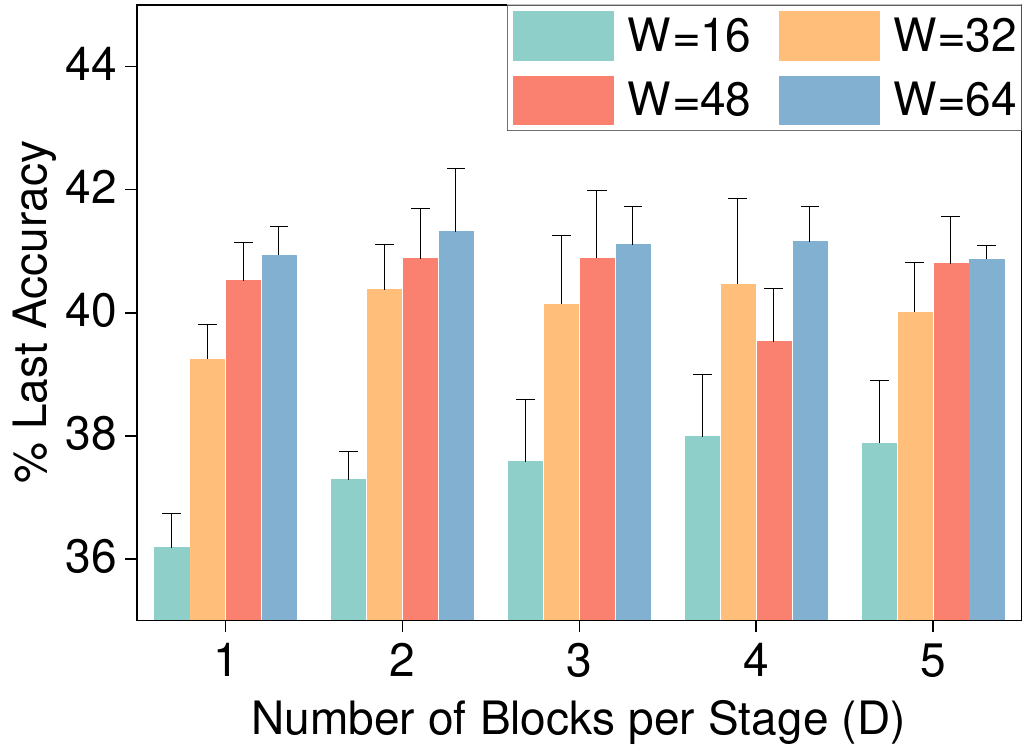}}
    \hfill
    \subcaptionbox{AIA in Class IL\label{subfig:aia_cil}}{\includegraphics[width = 0.23\textwidth]{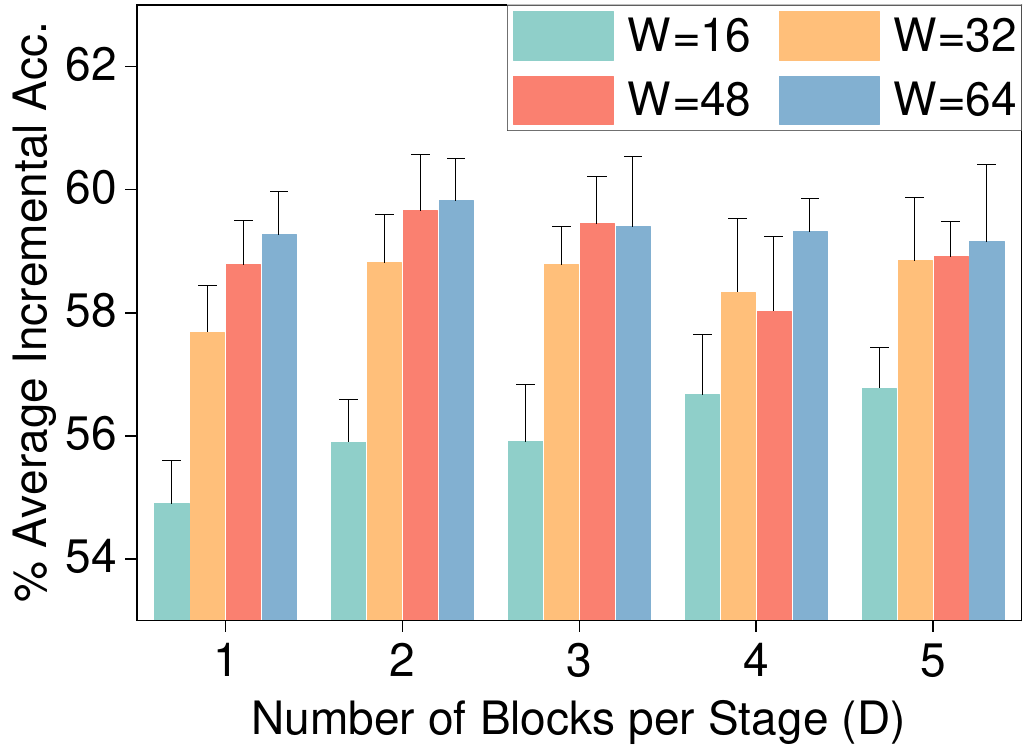}}
    \vspace{-2pt}
\caption{The performance of CL of ResNets with different network depth and width.}	
\label{fig:network_scaling}
\vspace{-2pt}
\end{figure}

Figure~\ref{fig:network_scaling} compares the CL performance of networks with different \textit{W} and \textit{D}. We observe that in both \textit{Task IL} (see Figure~\ref{subfig:la_til} and~\ref{subfig:aia_til}) and \textit{Class IL} (see Figure~\ref{subfig:la_cil} and~\ref{subfig:aia_cil}), the network with a larger width tends to exhibit better performance of CL in most cases. This phenomenon cannot be simply attributed to over-parameterization, as increasing depth does not yield similar results. In contrast, increasing the depth of the networks can even lead to obvious performance degradation in \textit{Task IL} (see Figure~\ref{subfig:la_til} and~\ref{subfig:aia_til}). These results indicate that wider and shallower networks may be more suitable for CL, explaining why ResNet-18 empirically shows better performance than ResNet-32 in Table~\ref{tab:network_components}. Therefore, we conclude that scaling the network depth and width appropriately is a potential way to enhance CL performance.

The above results raise a concern about whether the increase in the overall network width (i.e., initial width) or the width of the classifier (i.e., final width) contributes to the improvement of CL performance. To investigate this, we conduct further experiments on networks with the same final width but varying initial widths. In detail, we consider the network architectures with $D = 1$. To allow for adjustments to the final width, an additional convolutional layer with a 1 × 1 kernel size is inserted before the classifier for these networks. We maintain the final width of the network with $W = 64$ constant and adjust the output channels of the inserted convolutional layer for other networks to match it. The CL performance of networks with different initial and final widths are reported in Table~\ref{tab:initial_and_final_width}.

\begin{table}[ht]
    \centering
    \vspace{-2pt}
    \resizebox{.48\textwidth}{!}{
    \begin{tabular}{@{\hspace{2mm}}cc|cc|cc@{\hspace{2mm}}}
        \toprule
        \multirow{2.4}*{\makecell[c]{Initial\\Width}} &\multirow{2.4}*{\makecell[c]{Final\\Width}} &\multicolumn{2}{c|}{Performance in Task IL} &\multicolumn{2}{c}{Performance in Class IL}\\
        \cmidrule(lr){3-4} \cmidrule(lr){5-6}
        &&LA &AIA &LA &AIA\\
        \midrule
        
        \multirow{2}*{16} &64 &59.48{\scriptsize±2.03}	&65.65{\scriptsize±0.89} &35.82{\scriptsize±0.55}	&54.43{\scriptsize±0.47}\\
        &256 &71.51{\scriptsize±0.87} &73.89{\scriptsize±0.80} &36.16{\scriptsize±0.52} &55.18{\scriptsize±0.63}\\
        
        \midrule
        \multirow{2}*{32} &128 &68.24{\scriptsize±0.72}	&71.74{\scriptsize±0.69} &39.44{\scriptsize±0.81}	&57.83{\scriptsize±0.72}\\
        &256 &74.18{\scriptsize±0.94} &76.18{\scriptsize±0.46} &40.02{\scriptsize±0.74} &58.02{\scriptsize±0.87}\\
        
        \midrule
        \multirow{2}*{48} &192 &70.96{\scriptsize±1.70}	&74.94{\scriptsize±0.80} &40.52{\scriptsize±0.92}	&58.93{\scriptsize±0.77}\\
        &256 &73.12{\scriptsize±1.46} &76.64{\scriptsize±0.63} &40.53{\scriptsize±0.75} &59.08{\scriptsize±0.67}\\
        
        \midrule
        64 &256 &73.14{\scriptsize±1.51} &76.61{\scriptsize±0.66} &40.97{\scriptsize±0.56} &59.48{\scriptsize±0.85}\\
        \bottomrule
    \end{tabular}
    }
    \vspace{-2pt}
    \caption{
    The CL performance on different width configurations.
    }
    \label{tab:initial_and_final_width}
    \vspace{-2pt}
\end{table}

As shown in Table~\ref{tab:initial_and_final_width}, the results of ``performance in Task IL'' indicate that increasing the classifier width significantly benefits \textit{Task IL}. Moreover, networks with different initial widths but the same final width show comparable performance of CL. This implies that the classifier width, rather than the overall network width, is the main factor affecting the performance of wider networks in \textit{Task IL}. It also suggests that the adverse effect of GAP on \textit{Task IL} may stem from the reduction in the width of the classifier. By contrast, we observe from the results of ``performance in \textit{Class IL}'' that the impact of initial width on \textit{Class IL} is as essential as classifier width. This indicates that the overall network width and the width of the classifier both significantly affect \textit{Class IL}.

\subsection{ArchCraft Method}

To further investigate architectural designs that benefit CL, we propose ArchCraft, a straightforward yet powerful NAS method for CL. ArchCraft comprises a CL-friendly search space and employs a simple search strategy grounded in genetic algorithms.

\subsubsection{Search Space towards ArchCraft}

\textbf{What architectural design elements can be predetermined based on the aforementioned findings?} 
The above experiments show that ResNet can achieve good CL performance with slight modification, which demonstrates the great generalizability of its overall architecture for CL. Therefore, in our designed CL-friendly search space, the overall architecture of the networks is predetermined following the design of ResNet, i.e., each network architecture consists of a stem (i.e., a convolutional layer) and several units. Since the promising architecture is relatively shallow according to the experiments in Section~\ref{subsec:network_scaling}, we use a simple unit that comprises a single convolutional layer and a skip connection to explore more flexible architecture. More specifically, each convolutional layer is equipped with a 3 × 3 convolutional kernel and followed by batch normalization and ReLU. In addition, the network components are determined according to the corresponding optimal configurations summarized in Section~\ref{subsec:network_components} for \textit{Task IL} and \textit{Class IL}, respectively.

\textbf{What architectural design elements require further exploration?} To achieve a better CL performance, further exploration of certain architectural design elements is still necessary. First, although the significance of network width and depth for CL has been demonstrated, determining their optimal values necessitates further investigation. Second, the positions at which the channel number is increased and feature map down-sampling occurs are critical for CL performance, as they significantly influence the feature extraction capability of the network. However, their impact on CL has not been thoroughly examined in prior experiments. Therefore, the optimal designs of the locations of these operations are still necessary to search for. In summary, the architectural variations within our proposed search space encompass network width, depth, locations for increasing the channel number, and locations for down-sampling the feature map.

\textbf{How to concisely represent the candidate network architectures in search space?} We suggest an effective scheme to encode the architectures into a concise form that facilitates searching and adaptive scaling of parameters. The number of units and the initial number of channels are each denoted by a single code, which respectively represents the network depth and width. We specify that the size of the feature map can be halved zero to five times at chosen locations, which is indicated by five codes. Specifically, if the code $x$ is less than the number of units, it indicates that the size of the feature map is reduced by half through max pooling before the $(x+1)$-th unit; otherwise, it holds no significance. Similarly, the locations of the units in which the number of output channels is doubled are indicated by the other five codes. By the above means, each architecture in the proposed search space can be simply yet concisely encoded as a sequence comprising 12 codes. Additionally, the parameters of networks can be easily scaled by simultaneously reducing the network width and depth when the number of parameters exceeds the limit.

\subsubsection{Search Strategy towards ArchCraft}

In our evolutionary search strategy, the search process consists of an initialization phase and iterative evolutionary phases. During the initialization phase, we randomly generate a population of candidate architectures (i.e., individuals) within the defined search space. Subsequently, the fitness is evaluated for each individual. In each iteration of the evolutionary phase, the offspring is generated through the mutation after fitness evaluation. In detail, to generate an offspring individual, (i) two individuals are randomly selected from the population, and the one with higher fitness is chosen as the parent, (ii) the parent is copied and a randomly selected code of it is modified. In particular, when the code representing the number of units is modified, the codes that indicate the locations to down-sample feature maps and raise channel numbers are adjusted accordingly to keep the relative positions fixed. After the offspring generation, individuals are evaluated for fitness, and those with higher fitness are selected from the previous population and offspring to form the next population. Upon completion of the search process, the architecture exhibiting the highest fitness is chosen as the final design.

\subsubsection{Performance Evaluation towards ArchCraft}

We employ a direct method to evaluate the performance (i.e., fitness) of searched architectures. Specifically, each architecture is trained and assessed with the settings described in Section~\ref{sec:setup}, with the AIA serving as the fitness metric. 

\section{Evaluation of the Improved Architectures}

We apply the proposed ArchCraft method to design improved architectures for \textit{Task IL} and \textit{Class IL} separately, with a population size of 10 and an evolution of 20 generations. Then, we empirically compare these architectures with the baseline ones to demonstrate the efficacy of ArchCraft.

\subsection{Better CL Performance}

\subsubsection{Benchmark}

We choose CIFAR-100 and Imagenet-100 to evaluate the ArchCraft-guided architectures. Both datasets are divided into 20 tasks of 5 classes each and 10 tasks of 10 classes each to construct four benchmarks: C100-inc5, C100-inc10, I100-inc5, and I100-inc10. It should be highlighted that since the ArchCraft-guided architectures are crafted based on C100-inc5, the experiments conducted on C100-inc10 and ImageNet-100 serve to validate their generalizability.

\subsubsection{Evaluations in Task IL}

\textbf{AlexAC.} 
AlexNet~\cite{krizhevsky2012imagenet} is widely used as the basic architectures in \textit{Task IL}~\cite{serra2018overcoming,konishi2023parameter}. Thus, we craft two architectures with distinct parameter sizes based on them: \textbf{AlexAC-A} and \textbf{AlexAC-B} featuring marginally and far fewer parameters than AlexNet.

\textbf{Setup.} 
For CIFAR-100, we compare ArchCraft with several classical Task CL methods: SI~\cite{zenke2017continual}, HAT~\cite{serra2018overcoming}, WSN~\cite{kang2022forget}, and SPG~\cite{konishi2023parameter}. To make this comparison, we train AlexAC-A and AlexAC-B using a vanilla SGD optimizer, while AlexNet is trained with various CL methods. Additionally, we report the upper-bound performance for all networks. Note that the term `Upper-bound' refers to the result of training a separate model for each task, thereby eliminating any forgetting. Furthermore, we also provide a comparison between AlexNet and AlexAC-A on Imagenet-100, both of which are trained using a vanilla SGD optimizer.

\begin{table}[ht]
    \centering
    \vspace{-2pt}
    \resizebox{.48\textwidth}{!}{
    \begin{tabular}{@{\hspace{2mm}}ccccc@{\hspace{2mm}}}
        \toprule
        Network &\#P (M) &Method &C100-inc5 &C100-inc10\\
        \midrule
        \multirow{6.8}*{AlexNet} &\multirow{6.8}*{6.71} &Upper Bound &81.86 &73.07\\
        \cmidrule(lr){3-5}
        &&SGD &53.78 &56.92\\
        \cmidrule(lr){3-5}
        &&SI &70.3 &62.9\\
        &&HAT &71.8 &62.8\\
        &&SPG &75.9 &67.7\\
        &&WSN &76.9 &69.3\\
        \midrule
        \multirow{2.4}*{AlexAC-A} &\multirow{2.4}*{{6.28}\textcolor{forestgreen}{$\downarrow$ 6\%}} &Upper Bound &83.59 &75.09\\
        \cmidrule(lr){3-5}
        &&SGD &\textbf{82.38} \textcolor{red}{(+5.48)} &\textbf{73.91} \textcolor{red}{(+4.61)}\\
        \midrule
        \multirow{2.4}*{AlexAC-B} &\multirow{2.4}*{{0.92}\textcolor{forestgreen}{$\downarrow$ 86\%}} &Upper Bound &83.58 &74.72\\
        \cmidrule(lr){3-5}
        &&SGD &\underline{79.32} \textcolor{red}{(+2.42)} &\underline{70.87} \textcolor{red}{(+1.57)}\\
        \bottomrule
    \end{tabular}
    }
    \vspace{-2pt}
    \caption{The last accuracy of the AlexAC and AlexNet in \textit{Task IL}. `\#P' represents the number of parameters of the network used. \textbf{Bolded} indicates best performance. \underline{Underline} indicates second best.
    }
    \label{tab:result_til}
    \vspace{-2pt}
\end{table}

Table \ref{tab:result_til} details the comparison between the AlexAC series and AlexNet on CIFAR-100. It is evident that the AlexAC series consistently outperforms AlexNet in all cases. In particular, when both are trained using the vanilla SGD, AlexAC-A achieves 28.6\% and 16.99\% higher LA than the AlexNet on C100-inc5 and C100-inc10 settings, respectively, while utilizing fewer parameters. The magnitude of this improvement surpasses that achieved by state-of-the-art methods (28.6\% and 16.99\% vs. 23.12\% and 12.38\%). These results underscore the pivotal role of network architectures in CL. This implies that advancements in CL can be propelled not only through learning methods but also through improved architectural designs. Moreover, AlexAC-B attains superior performance compared to state-of-the-art methods with 86\% fewer parameters. In conclusion, these observations strongly emphasize the significance of enhanced architecture in CL. 

\begin{table}[ht]
    \centering
    \vspace{-2pt}
    \resizebox{.44\textwidth}{!}{
    \begin{tabular}{@{\hspace{2mm}}ccccc@{\hspace{2mm}}}
        \toprule
        Network &\#P (M) &Method &I100-inc5 &I100-inc10\\
        \midrule
        AlexNet &6.71 &SGD  &35.36 &33.38\\
        AlexAC-A &{6.28}\textcolor{forestgreen}{$\downarrow$ 6\%} &SGD &\textbf{52.02} &\textbf{45.86}\\
        \bottomrule
    \end{tabular}
    }
    \vspace{-2pt}
    \caption{The last accuracy on Imagenet-100 (I100) in \textit{Task IL}.
    }
    \label{tab:result_til_i100}
    \vspace{-2pt}
\end{table}

Table \ref{tab:result_til_i100} reports the comparison between AlexAC-A and AlexNet on Imagenet-100. We observe that AlexAC-A also achieves significantly higher LA (16.66\% and 12.48\%) with fewer parameters compared to AlexNet. This result demonstrates that ArchCraft can design generally useful architectures that facilitate \textit{Task IL}.

\begin{table*}[t]
    \centering
    \resizebox{.96\textwidth}{!}{
    \begin{tabular}{@{\hspace{2mm}}ccccccccccc|cc@{\hspace{2mm}}}
        \toprule
        \multirow{2.4}*{Method} &\multirow{2.4}*{Network} &\multirow{2.4}*{\#P (M)} &\multicolumn{2}{c}{C100-inc5} &\multicolumn{2}{c}{C100-inc10} &\multicolumn{2}{c}{I100-inc5} &\multicolumn{2}{c|}{I100-inc10} &\multicolumn{2}{c}{Max Improvement}\\
        \cmidrule(lr){4-5} \cmidrule(lr){6-7} \cmidrule(lr){8-9} \cmidrule(lr){10-11} \cmidrule(lr){12-13}
        &&&LA &AIA &LA &AIA &LA &AIA &LA &AIA &LA &AIA\\
        \midrule
        \multirow{4.4}*{Replay} &ResNet-32 &0.46 &39.10 &58.17 &40.02 &58.21 &- &- &- &- &- &-\\
        &ResAC-B &0.44(\textcolor{forestgreen}{$\downarrow$ 4\%}) &\textbf{40.45} &\textbf{59.67} &\textbf{42.79} &\textbf{59.99} &- &- &- &- &\textcolor{red}{+2.77} &\textcolor{red}{+1.78}\\
        \cmidrule(lr){2-13}
        &ResNet-18 &11.17 &40.04 &58.80 &43.23 &60.42 &36.30 &57.30 &41.00 &59.21 &- &-\\
        &ResAC-A &8.63\textcolor{forestgreen}{$\downarrow$ 23\%} &\textbf{42.99} &\textbf{62.52} &\textbf{46.62} &\textbf{63.36} &\textbf{36.78} &\textbf{57.40} &\textbf{42.44} &\textbf{60.07} &\textcolor{red}{+3.39} &\textcolor{red}{+3.72}\\
        \midrule
        \multirow{4.4}*{iCaRL}  &ResNet-32 &0.46 &46.67 &63.47 &48.80 &64.18 &- &- &- &- &- &-\\
        &ResAC-B &0.44\textcolor{forestgreen}{$\downarrow$ 4\%} &\textbf{47.94} &\textbf{64.17} &\textbf{50.11} &\textbf{64.42} &- &- &- &- &\textcolor{red}{+1.31} &\textcolor{red}{+0.70}\\
        \cmidrule(lr){2-13}
        &ResNet-18 &11.17 &47.32 &64.13 &52.77 &66.04 &44.10 &62.36 &50.98 &67.11 &- &-\\
        &ResAC-A &8.63\textcolor{forestgreen}{$\downarrow$ 23\%} &\textbf{52.6} &\textbf{68.71} &\textbf{55.52} &\textbf{69.62} &\textbf{45.12} &\textbf{63.98} &\textbf{52.46} &\textbf{68.42} &\textcolor{red}{+5.28} &\textcolor{red}{+4.58}\\
        \midrule
        \multirow{4.4}*{WA}  &ResNet-32 &0.46 &46.95 &62.93 &53.35 &66.61 &- &-&- &- &- &-\\
        &ResAC-B &0.44\textcolor{forestgreen}{$\downarrow$ 4\%} &\textbf{51.31} &\textbf{66.39} &\textbf{54.89} &\textbf{67.73} &- &- &- &- &\textcolor{red}{+4.36} &\textcolor{red}{+3.46}\\
        \cmidrule(lr){2-13}
        &ResNet-18 &11.17 &45.11 &62.06 &56.59 &68.89 &46.06 &62.96 &55.04 &68.60 &- &-\\
        &ResAC-A &8.63\textcolor{forestgreen}{$\downarrow$ 23\%} &\textbf{53.23} &\textbf{69.19} &\textbf{59.79} &\textbf{71.40} &\textbf{49.94} &\textbf{67.20} &\textbf{58.86} &\textbf{71.56} &\textcolor{red}{+8.12} &\textcolor{red}{+7.13}\\ 
        \midrule
        \multirow{4.4}*{Foster}  &ResNet-32 &0.46 &47.78 &62.36 &54.36 &67.14 &- &- &- &- &- &-\\
        &ResAC-B &0.44\textcolor{forestgreen}{$\downarrow$ 4\%} &\textbf{53.50} &\textbf{67.34} &\textbf{58.17} &\textbf{69.44} &- &- &- &- &\textcolor{red}{+5.72} &\textcolor{red}{+4.98}\\
        \cmidrule(lr){2-13}
        &ResNet-18 &11.17 &49.03 &61.97 &55.98 &68.38 &53.26 &65.20 &60.58 &69.36 &- &-\\
        &ResAC-A &8.63\textcolor{forestgreen}{$\downarrow$ 23\%} &\textbf{57.22} &\textbf{69.99} &\textbf{61.44} &\textbf{72.54} &\textbf{54.32} &\textbf{66.41} &\textbf{61.94} &\textbf{71.16} &\textcolor{red}{+8.19} &\textcolor{red}{+8.02}\\ 
        \bottomrule
    \end{tabular}
    }
    \caption{The CL performance of ArchCraft in \textit{Class IL}. `\#P' represents the number of parameters of the network used.}
    \label{tab:result_cil}
\end{table*}

\subsubsection{Evaluations in Class IL}

\textbf{ResAC.} 
ResNet series are widely used as the basic architectures in \textit{Class IL}. Thus, we craft two architectures with distinct parameter sizes based on them: \textbf{ResAC-A} and \textbf{ResAC-B} featuring fewer parameters than ResNet-18 and ResNet-32.

\textbf{Setup.} 
We evaluate our method combined with several classical Class CL methods: Replay, iCaRL~\cite{rebuffi2017icarl}, WA~\cite{zhao2020maintaining}, and Foster~\cite{wang2022foster}. Hyper-parameters for all methods adhere to the settings in the open-source library~\cite{zhou2023pycil}, and a fixed memory size of 2,000 exemplars is utilized for all methods.

Table~\ref{tab:result_cil} reports the LA and AIA of the ResAC series and ResNet series in \textit{Class IL}. In this case, ResAC-A consistently outperforms ResNet-18 across all methods, datasets, and incremental steps. In particular, ResAC-A achieves a maximum 8.19\% higher LA and 8.02\% higher AIA with 23\% fewer parameters than the ResNet-18. Moreover, ResAC-A demonstrates superior performance on ImageNet, indicating robust generalizability. This suggests that the enhancement afforded by the improved network architecture can effectively generalize to various methods and settings. Furthermore, under stricter parameter constraints, ResAC-B, compared to ResNet-32, attains up to 5.72\% and 4.98\% improvements in LA and AIA with 4\% fewer parameters. In conclusion, the proposed method can serve as a valuable complement to CL.

\subsection{Reducing the Number of Parameters}

In this subsection, we thoroughly examine the effects of parameter sizes and network architecture. To simplify, we choose one typical method Foster as a baseline.

\begin{figure}[ht]
    \centering
    
    \subcaptionbox{\label{subfig:paci}}{\includegraphics[width = 0.23\textwidth]{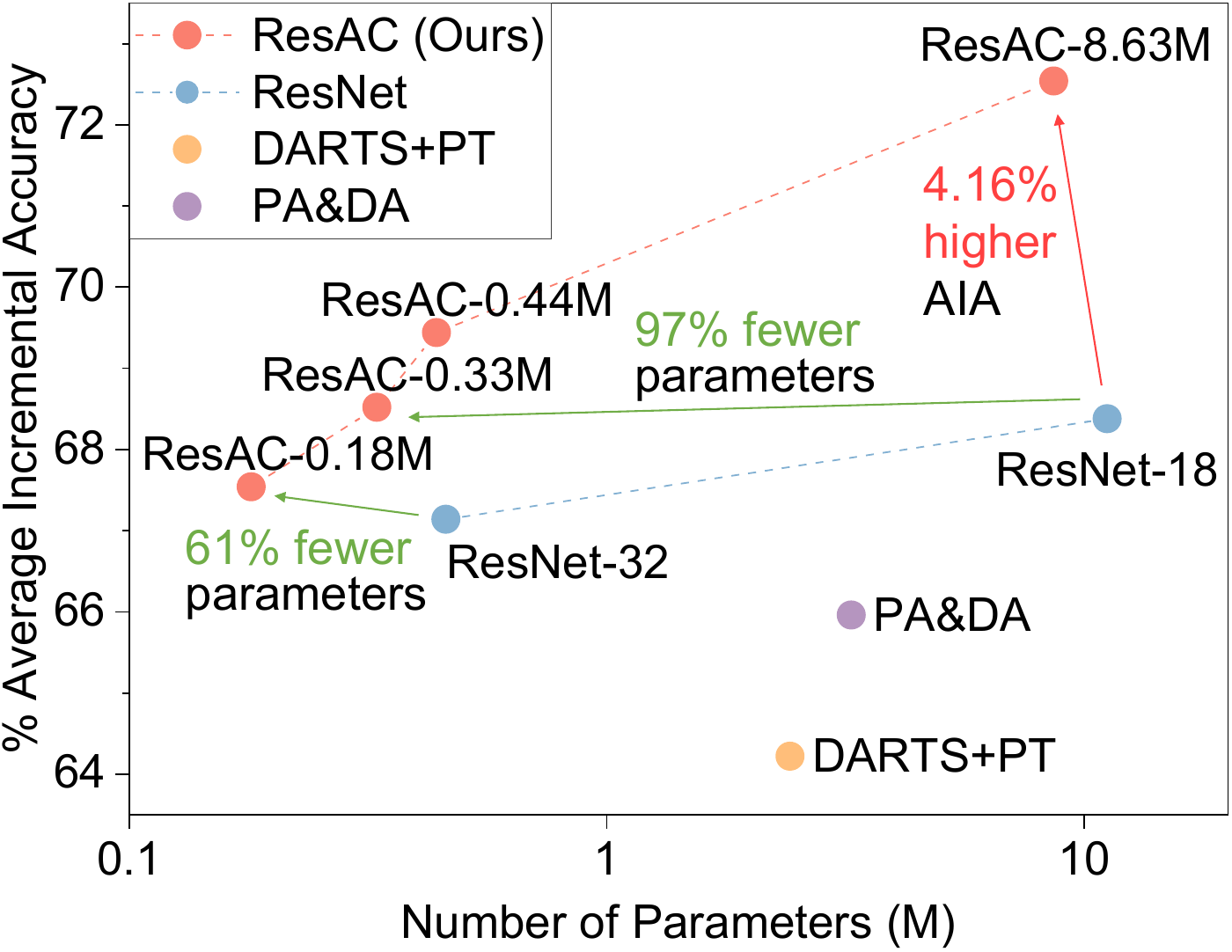}}
    \hfill
    \subcaptionbox{\label{subfig:paim}}{\includegraphics[width = 0.23\textwidth]{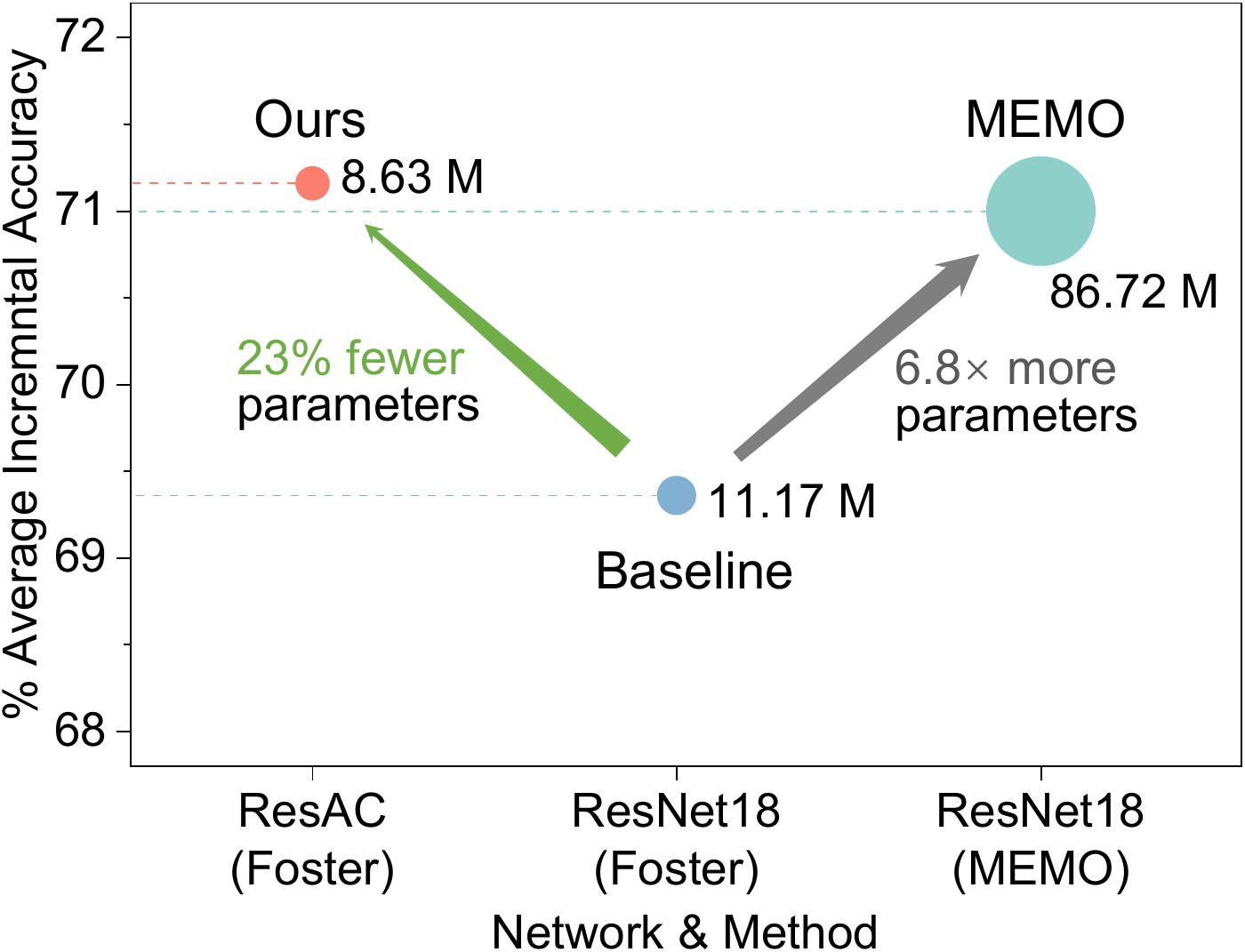}}

\vspace{-2pt}
\caption{(a) \textbf{Performance of CL \textit{vs.} Number of Parameters} on C100-inc10. 
(b) 
The comparison of ArchCraft and expansion-based method (i.e., MEMO) in terms of parameters on I100-inc10. 
}
\label{fig:params}
\vspace{-6pt}
\end{figure}

First, Figure~\ref{fig:params}a yields three conclusions: (i) ArchCraft can craft neural networks with controllable parameter sizes, e.g., from ResAC-0.18M to ResAC-8.63M.  (ii) ResAC series achieve better CL performance with significantly fewer parameters than the naive ResNet series, e.g., with similar performance, 61\% and 97\% fewer parameter sizes (from 0.46M to 0.18M and from 11.17M to 0.33M, respectively); With similar parameter sizes, 2.3\% higher performance on AIA  (from 67.14\% to 69.44\%), or more.  (iii) Existing state-of-the-art NAS methods, e.g., DARTS+PT~\cite{wang2021rethinking} and PA\&DA~\cite{lu2023pa}, fail to design CL-friendly architectures. For this point, we think that most existing NAS methods can not adaptively adjust the parameter sizes, since the key architectural elements (i.e., depth, width) that determine the number of parameters rely on predefined manually in their search space~\cite{liu2018darts,wan2022redundancy}.  In summary, ArchCraft recrafts AlexNet/ResNet into AlexAC/ResAC to guide a well-designed network architecture for CL with fewer parameter sizes.

In addition, as Figure~\ref{fig:params}b, we compare the proposed method with the expansion-based method MEMO~\cite{zhou2022model}. As is known to all, this category usually obtains performance improvement at the expense of increasing the parameter size of the baseline (from 11.17M to 86.72M). This leads to the limited availability of methods of this category. In radical contrast, ArchCraft achieves a better result with much fewer parameters than the baseline architecture (from 11.17M to 8.63M). To sum up, these empirical results provide fresh insights for advancing CL.

\subsection{Stronger Stability and Plasticity}

\begin{figure}[ht]
    \centering
    
    \subcaptionbox{AF in Task IL}{\includegraphics[width = 0.23\textwidth]{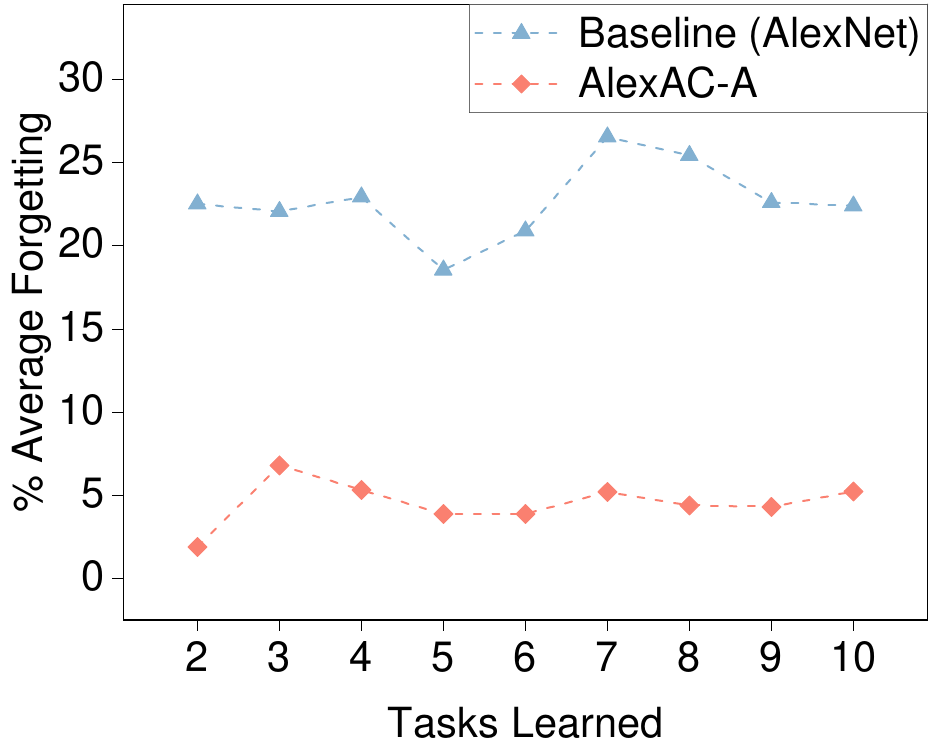}}
    \hfill
    \subcaptionbox{New Task Acc. in Task IL}{\includegraphics[width = 0.23\textwidth]{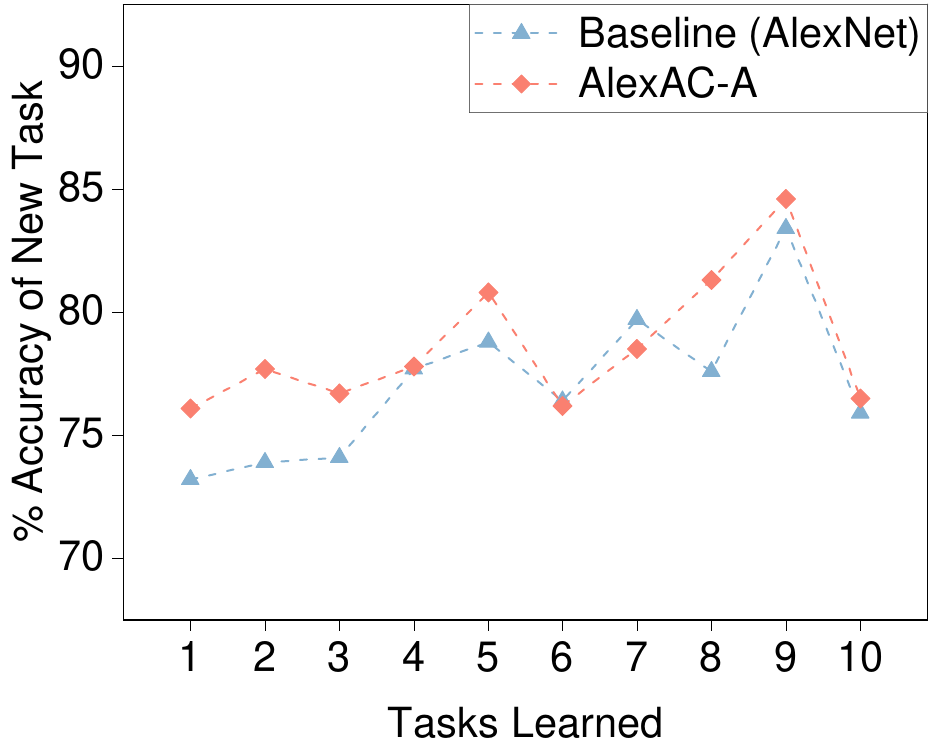}}
    \\
    \subcaptionbox{AF in Class IL}{\includegraphics[width = 0.23\textwidth]{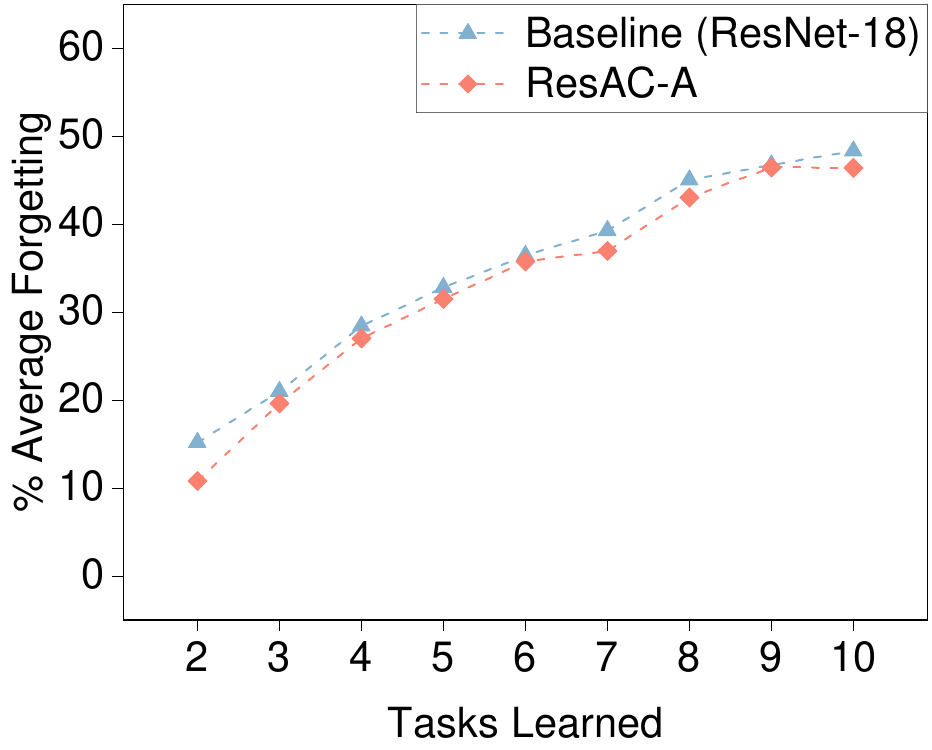}}
    \hfill
    \subcaptionbox{New Task Acc. in Class IL}{\includegraphics[width = 0.23\textwidth]{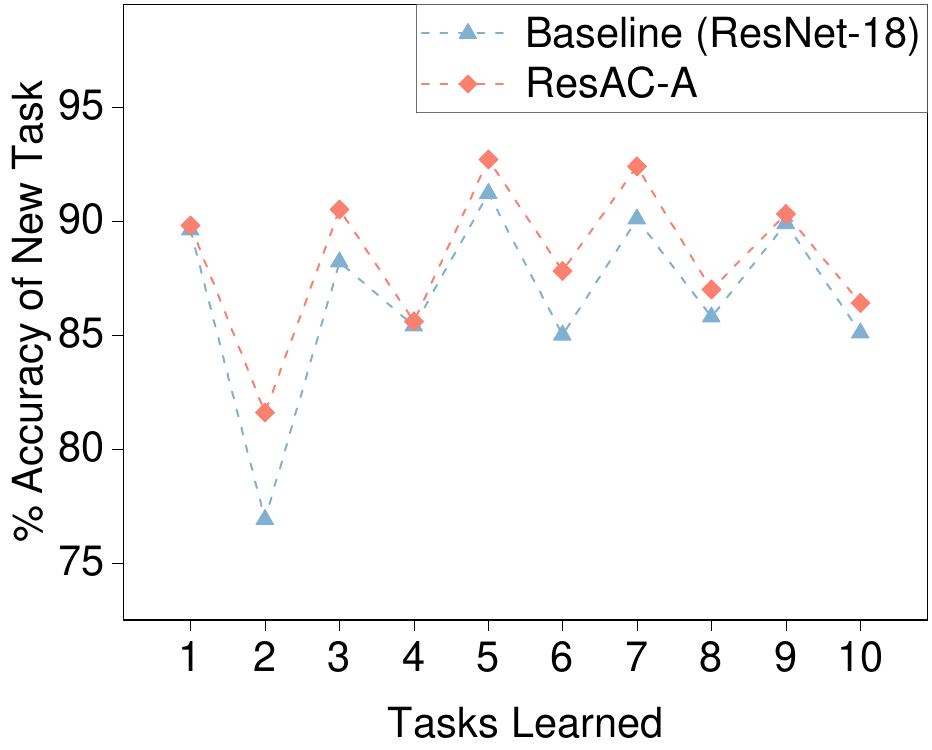}}
    \vspace{-2pt}
\caption{The average forgetting on the previous tasks and performance on the current task of the ArchCraft-guided networks and the baseline ones in \textit{Task IL} and \textit{Class IL}. 
}	
\label{fig:af_and_newAcc}
\vspace{-2pt}
\end{figure}

To further scrutinize the effectiveness of ArchCraft, we report the average forgetting and accuracy of the new task on C100-inc10 for \textit{Task IL} and \textit{Class IL}. As illustrated in Figure~\ref{fig:af_and_newAcc}, the ArchCraft-guided architectures exhibit less forgetting on the previous task and higher accuracy on the current task than the baselines in a consistent setting. This observation suggests that stability and plasticity can be concurrently improved by employing architectures designed by ArchCraft. Furthermore, it emphasizes the critical role of heightened stability in enhancing CL performance.

\textbf{What contributes to the stronger stability of the ArchCraft-guided network architecture?} To investigate this, we quantitatively assess the similarity of network representations across models from all incremental stages during CL. In detail, we utilize Centered Kernel Alignment (CKA)~\cite{kornblith2019similarity}, a robust and widely used metric, to compute the similarity between representations. To facilitate this comparison, we extract feature maps from the final convolutional layer of each model by inputting the same set of instances. Subsequently, CKA is applied to measure the similarity between these feature maps.

\begin{figure}[ht]
    \centering
    
    \subcaptionbox{AlexNet in Task IL}{\includegraphics[width = 0.23\textwidth]{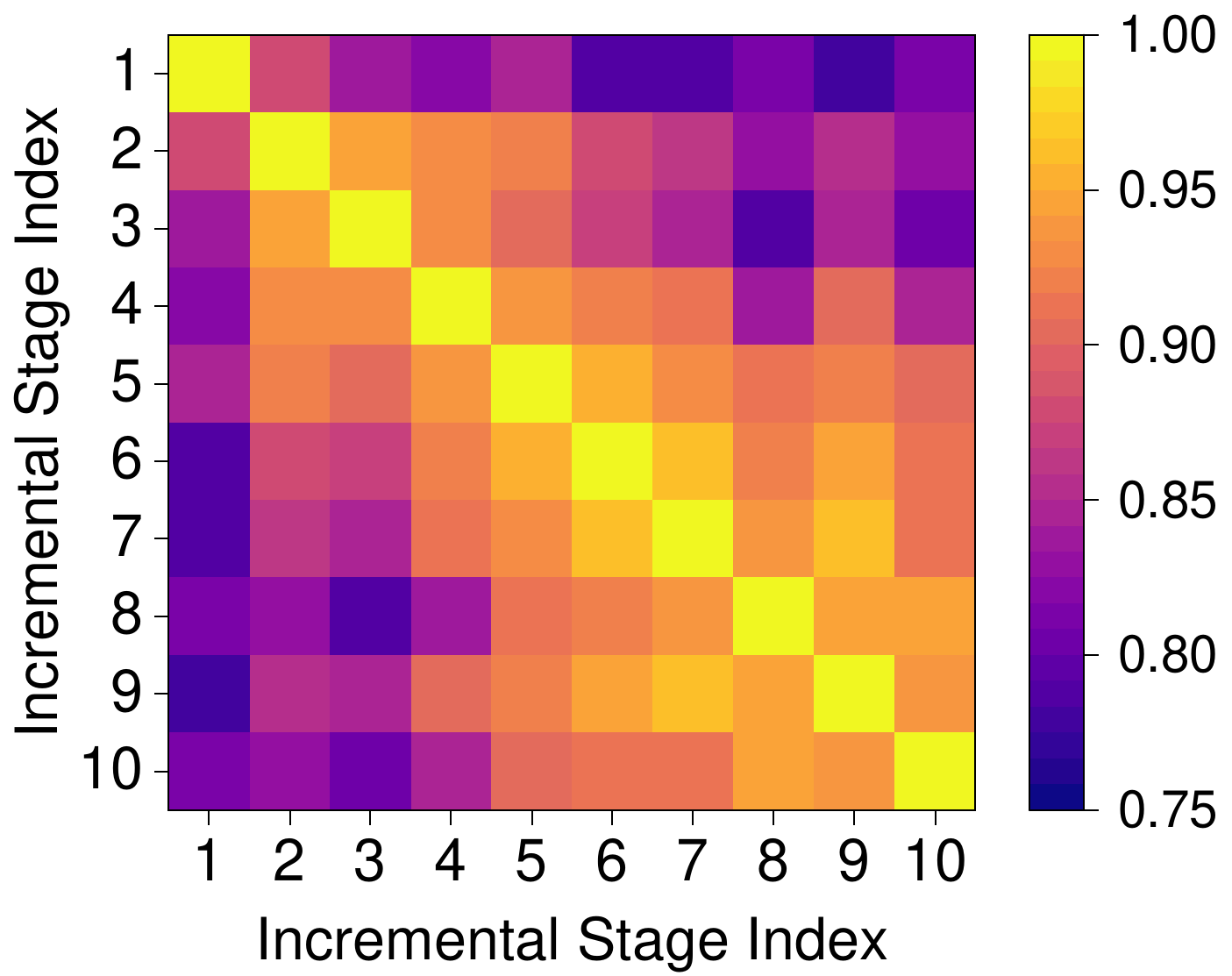}}
    \hfill
    \subcaptionbox{AlexAC-A in Task IL}{\includegraphics[width = 0.23\textwidth]{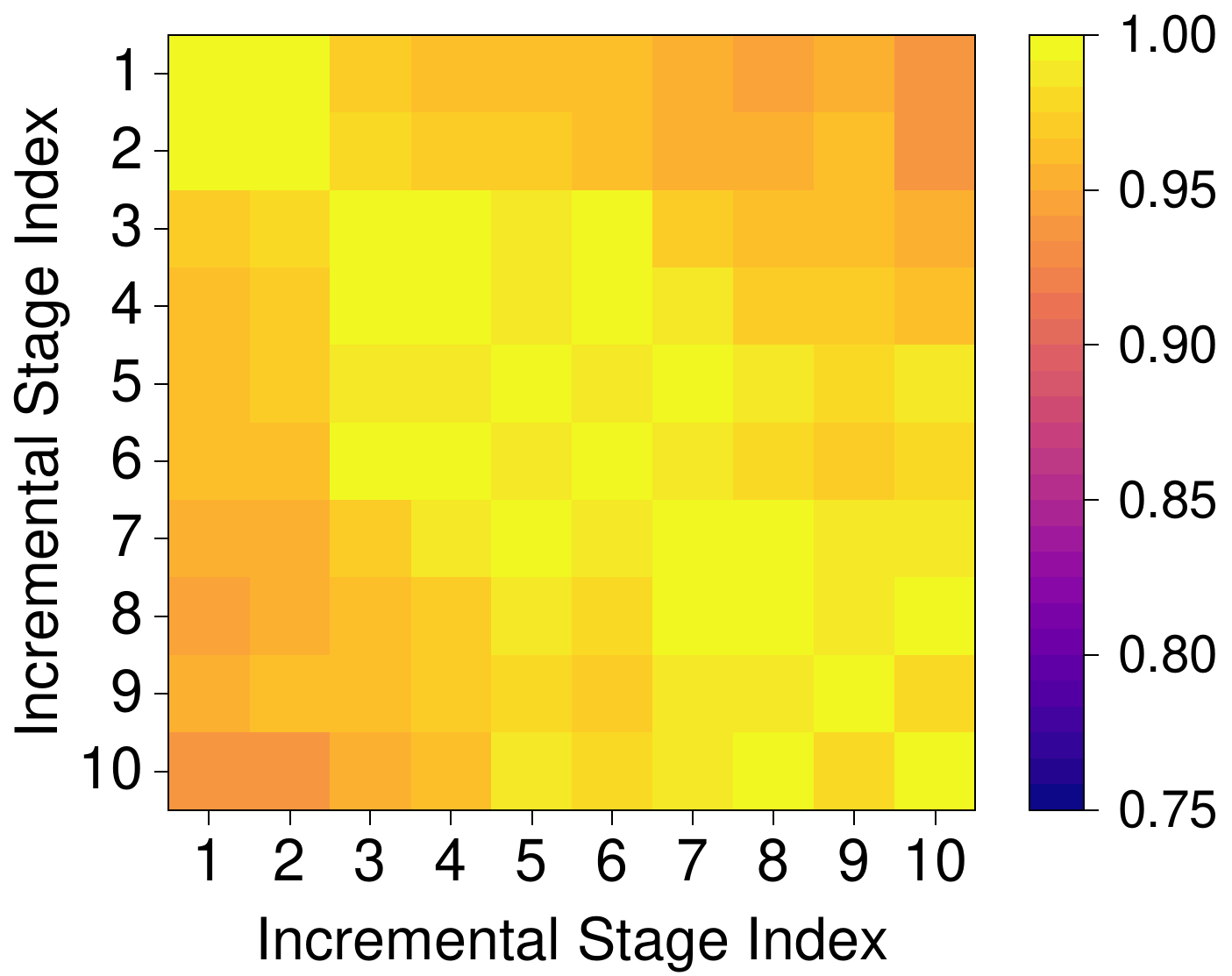}}
    \\
    \subcaptionbox{ResNet in Class IL}{\includegraphics[width = 0.23\textwidth]{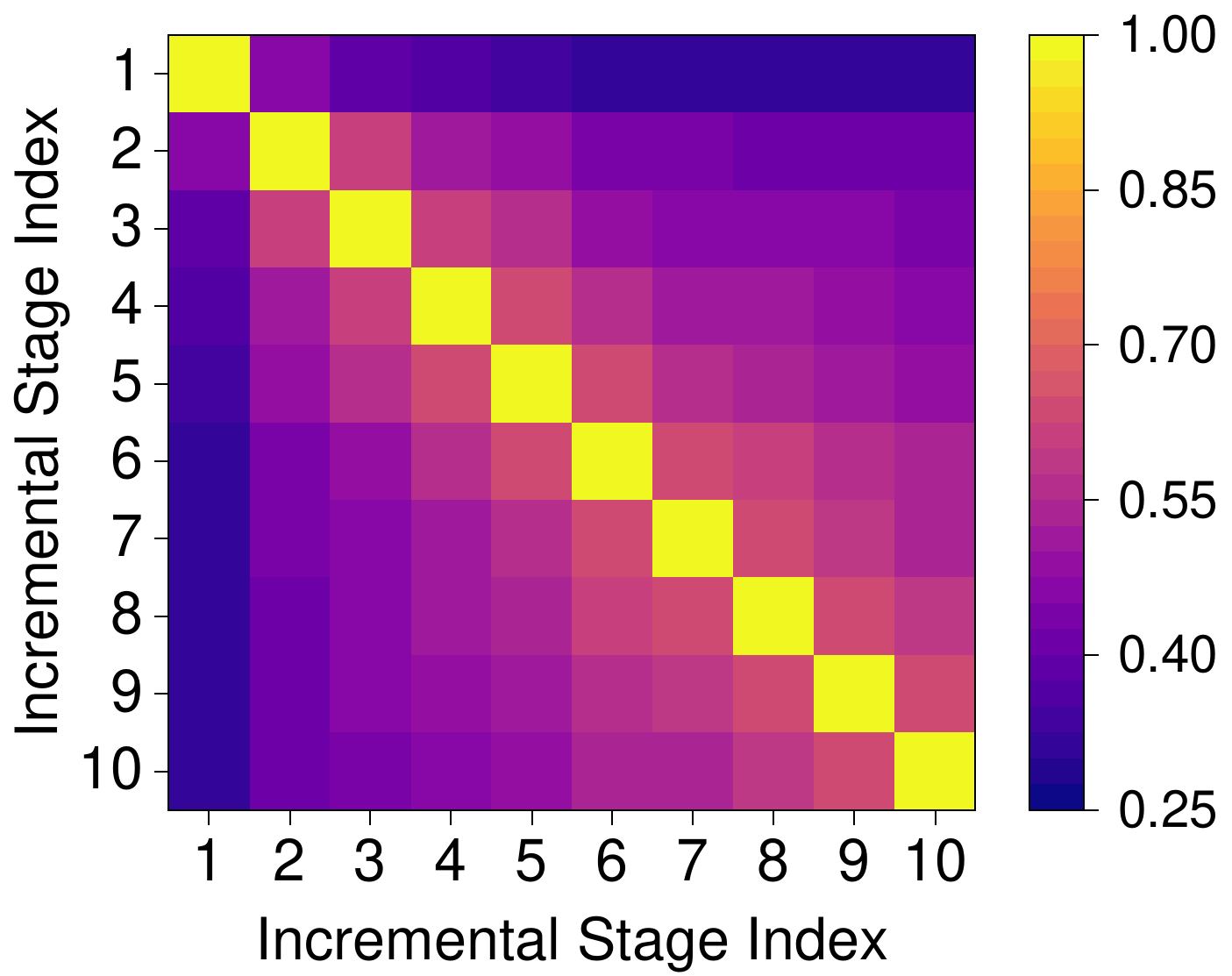}}
    \hfill
    \subcaptionbox{ResAC-A in Class IL}{\includegraphics[width = 0.23\textwidth]{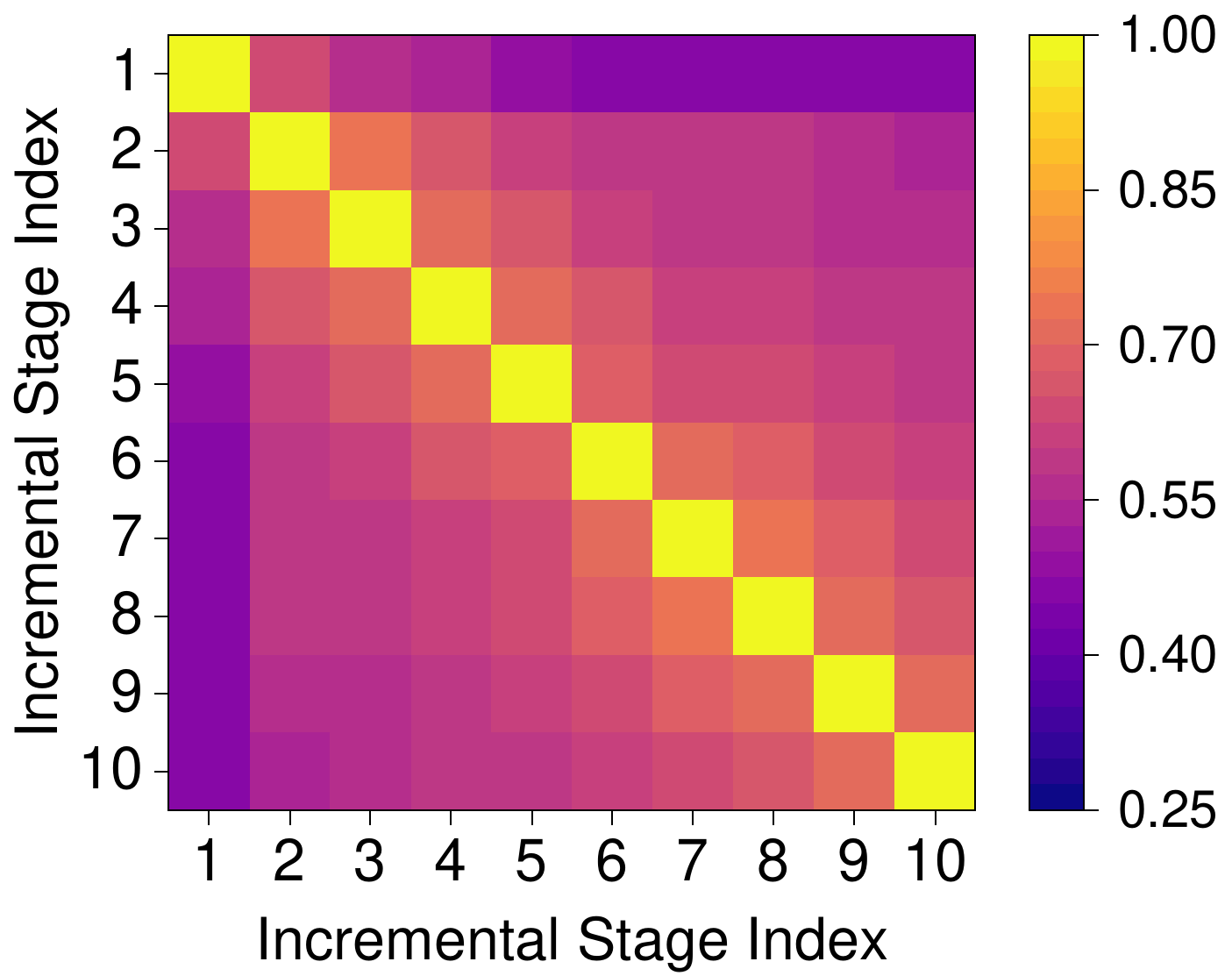}}
    \vspace{-2pt}
\caption{CKA visualization of the ArchCraft-guided networks and the baseline ones in different incremental stages across.}	
\label{fig:cka}
\vspace{-2pt}
\end{figure}

To put it more intuitively, we present the similarity matrix in Figure~\ref{fig:cka}. Figure~\ref{fig:cka} illustrates that features extracted by the ArchCraft-guided network architecture show a higher degree of similarity across all incremental stages compared to the baseline ones. This observation suggests that the network architecture guided by ArchCraft more effectively extracts shared features between incremental tasks, facilitating adaptation to new tasks with minimal changes. Thus, ArchCraft results in better overall stability while maintaining plasticity.

\section{Conclusion}

This paper identifies specific architectural designs that influence CL. The proposed ArchCraft bridges the gap between network architecture design and CL, achieving superior CL performance while utilizing a significantly more compact number of parameters than naive CL architectures. We affirm that ArchCraft holds immediate practical relevance in the design of CL-friendly networks. We hope this work inspires further study on overcoming catastrophic forgetting and enhancing CL performance from the architectural perspective.

\section*{Acknowledgements}

This work was supported by National Natural Science Foundation of China under Grant 62276175.

\bibliographystyle{named}
\bibliography{ijcai24}

\end{document}